\documentclass[letterpaper,twocolumn]{IEEEAerospaceCLS}  

\usepackage{graphicx}    
\usepackage[numbers]{natbib}
\usepackage[bookmarks=true]{hyperref} 
\usepackage[short]{optidef}           
\usepackage[ruled]{algorithm2e}       
\usepackage{pgfplots}                 
\usepackage{tikzscale}                
\usepackage{stmaryrd}

\newcommand{\ignore}[1]{}  

\pgfplotsset{compat=1.16}

\begin{document}

\title{Robust Entry Vehicle Guidance with Sampling-Based Invariant Funnels}


\author{
Remy Derollez\\ 
Stanford University\\
Department of Aeronautics and Astronautics \\
496 Lomita Mall\\
Stanford CA 94305\\
derollez@stanford.edu
\and 
Simon Le Cleac'h\\ 
Stanford University\\
Department of Mechanical Engineering \\
496 Lomita Mall\\
Stanford CA 94305\\
simonlc@stanford.edu
\and 
Zachary Manchester \\
Carnegie Mellon University \\
Robotics Institute \\
5000 Forbes Ave\\
Pittsburgh, PA 15213\\
zacm@cmu.edu \thanks{\footnotesize 978-1-7281-7436-5/21/$\$31.00$ \copyright2021 IEEE}}

\maketitle 


\thispagestyle{plain}
\pagestyle{plain}

\begin{abstract}
    Managing uncertainty is a fundamental and critical issue in spacecraft entry guidance. This paper presents a novel approach for uncertainty propagation during entry, descent and landing that relies on a new sum-of-squares robust verification technique. Unlike risk-based and probabilistic approaches, our technique does not rely on any probabilistic assumptions. It uses a set-based description to bound uncertainties and disturbances like vehicle and atmospheric parameters and winds. The approach leverages a recently developed sampling-based version of sum-of-squares programming to compute regions of finite time invariance, commonly referred to as ``invariant funnels''. We apply this approach to a three-degree-of-freedom entry vehicle model and test it using a Mars Science Laboratory reference trajectory. We compute tight approximations of robust invariant funnels that are guaranteed to reach a goal region with increased landing accuracy while respecting realistic thermal constraints.
\end{abstract}
\vspace{-8mm}

\tableofcontents

\begin{figure}[t]
	\centering
	\includegraphics[width=85mm]{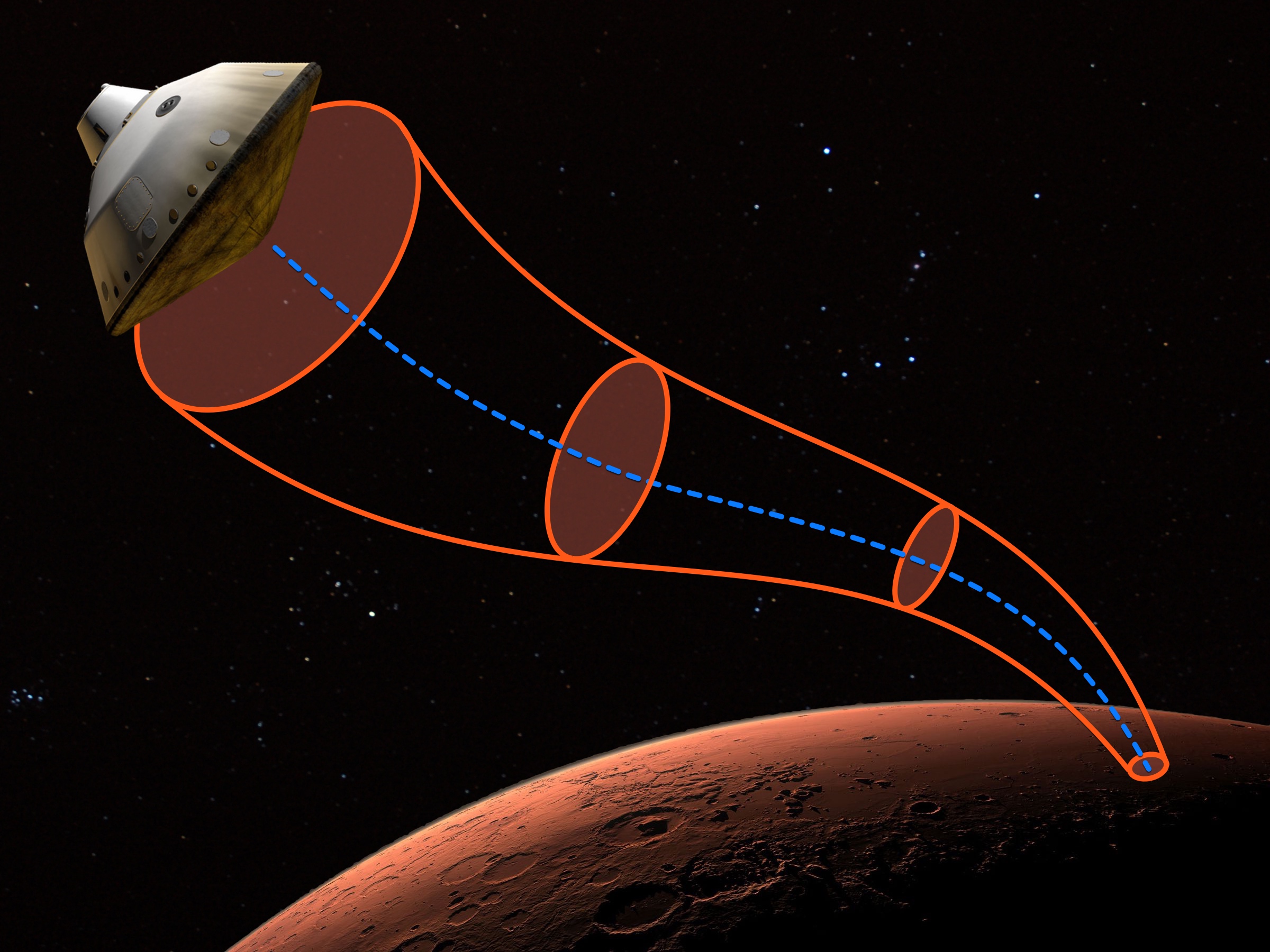}
	\caption{Mars entry funnel concept. The funnel provides conservative bounds on the maximum deviation of the closed-loop system from the nominal trajectory in the presence of disturbances and model uncertainty.
	\vspace{-8mm}
	}
	\label{fig:funnel-concept}
\end{figure}

\section{Introduction}


The future ambitions of multiple space agencies, including NASA, target the Moon and Mars as future destinations. Robotic missions such as Mars 2020 extend our knowledge of the red planet incrementally by performing increasingly complex scientific experiments and in-situ exploration tasks. Ultimately, robotic missions also pave the way for crewed missions to Mars that are studied and planned both by national space agencies and private companies. The Entry Descent and Landing (EDL) phase is a crucial part of all missions that require landing on another celestial body.

Past missions to Mars were primarily concerned with safe landing, paying little respect to the value of the landing site \cite{Li2014}. As ambitions grow, mission and specifically landing requirements reach new levels. New missions seek to land larger and more complex payloads on the surface of Mars. Improved landing accuracy is also desired, whether it is to achieve pinpoint landing at a scientifically relevant site, land near pre-positioned assets, or ensure the proximity to an established human colony on Mars. It is currently envisioned that future crewed missions to our neighbour planet will likely require safe landing of twenty metric tons mass with an accuracy better than one hundred meters.

Landing on Mars is a hard problem that encompasses major challenges \cite{Braun2006}. Mars' atmosphere is thin enough to affect vehicles thermally \cite{tauber1991stagnation} in a critical way, but does not provide the same braking as Earth's atmosphere. This situation still prevents missions landing at higher altitude points. In addition, knowledge of the planetary environment, atmospheric parameters, and winds are all imperfect. Uncertainties on position, velocity, and aerodynamic properties of the spacecraft also still represent a substantial issue \cite{Bose2013}.

\vspace{1mm}
Many of the entry guidance strategies currently in use still rely heavily on methods developed for the Viking missions in the 70's \cite{Braun2006, zang2010overview}. Until recently, most landers had flown an unguided ballistic atmospheric entry. An integrated closed-loop guidance and control strategy was used for the Mars Science Laboratory (MSL) and Curiosity missions. Recently algorithms and approaches leveraging optimization, optimal control, and adaptive techniques \cite{Roenneke2001} have emerged to address the entry trajectory optimization, guidance, and control problem. Predictor-corrector methods have been developed and tested for better performance \cite{Li2014}. Li uses direct collocation and nonlinear programming to achieve increased precision during entry \cite{Li2011}. The work of Benito expands on landing footprints \cite{Saraf2004}  using the concepts of reachable and controllable sets \cite{benito2010reachable}. Particle-swarm optimization approaches have been used to tackle the re-entry problem \cite{Rahimi2013}. Steinfeldt's work leverages the concept of a state-dependent Riccati equation and applies it to atmospheric entry guidance \cite{Steinfeldt2010}. Finally, machine learning and neural networks have also begun to be applied to the topic \cite{Li2015}.

On top of traditional entry guidance techniques, uncertainty analysis \cite{luo2017review} is usually performed using Monte Carlo methods \cite{lockwood2001entry}, which are computationally expensive. Polynomial Chaos Expansion (PCE) \cite{jones2013nonlinear} methods have also been applied to hypersonic flight dynamics \cite{Prabhakar2010, Derollez}, but do not scale well to the higher dimensions needed to model realistic problems. More recently, Jin and al. applied linear covariance techniques to the entry problem on a three-degree-of-freedom model and compared the results to a Monte Carlo analysis \cite{jin2018development}. Woffinden and al. went further to apply similar linear covariance analysis techniques to atmospheric flight during the EDL phase, but this time on a full six-degree-of-freedom dynamical model to generate navigation system requirements for the Safe and Precise Landing – Integrated Capabilities Evolution (SPLICE) project \cite{Woffinden2019}. 
While these new approaches are usually faster than traditional Monte Carlo analyses, some require strong assumptions on the probability distributions of the state and/or parameters, which are difficult to infer given our limited knowledge of the Mars environment. Many techniques are also difficult to integrate in an optimization framework. Following those observations, the authors previously investigated a non-probabilistic, scalable alternative using sampling and convex optimization \cite{Derollez}.

In parallel, robust motion planning tools and systems verification techniques have been developed in the robotics community with the goal to obtain strong guarantees on performance and safety of robotic systems. In particular, some set-based uncertainty propagation and verification approaches rely on Sum-Of-Squares (SOS) programming for the computation of regions around nominal trajectories in which the system under study is guaranteed to be at all time given some bounded sets for the uncertainties and disturbances. These regions are referred to as regions of finite-time invariance (or ``invariant funnels'') \cite{Tedrake2010, Majumdar2017, Tobenkin2010}. Even if we cannot deny the power of these approaches, their tractability has usually been an issue for application to real-life systems.

It is the authors' belief that Mars entry trajectory planning and robust uncertainty propagation are important prerequisites for
the implementation and integration of relevant guidance and control strategies. Improvement in these domains should allow us to reach the safety and accuracy goals driven by future missions' requirements. In that context, this paper presents a novel approach for uncertainty propagation during EDL that relies on a recently developed sum-of-squares robust verification technique. Combining the recent work from Shen and Tedrake \cite{Shen} and the concept of invariant funnels, a sampling-based SOS approach is leveraged to compute regions of finite-time invariance for the Mars entry guidance problem. The resulting technique does not rely on any probabilistic assumptions. The funnels we compute provide strong guarantees on safety and accuracy, while substantially reducing computational burden. Figure \ref{fig:funnel-concept} shows a conceptual view of the entry funnel.

Our primary contributions include: 
\begin{itemize}
    \item The introduction of sum-of-squares verification techniques for robust motion planning and uncertainty propagation to EDL applications.
    \vspace{+2mm}
    \item The presentation of a novel sampling-based SOS funnel computation approach providing safety and performance guarantees.
    \vspace{+2mm}
    \item The application of this new technique to the Mars entry guidance problem and the demonstration of its performance compared to traditional methods.
    \vspace{+2mm}
    \item The development of the open-source SystemVerification.jl software toolbox to compute invariant funnels for dynamical systems using both traditional and sampling-based approaches.
\end{itemize}

This paper is organized as follows: Section \ref{section:backrgound} provides some background on sum-of-squares programming and regions of finite-time invariance. Using these concepts, Section \ref{section:sampling-based_sos_funnel} details the sampling-based SOS approach for funnel computation at the core of this work. Section \ref{section:simulation_results} presents the results of the approach applied to the Mars entry guidance problem. Finally Section \ref{section:discussion} provides some analysis on the results and ends with some ways forward.

\vspace{-3mm}
\section{Background}
\label{section:backrgound}

\subsection{Sum-of-Squares Programming}

Sum-of-squares methods have been developed in the robotics community over the past decade to enable control of complex mechanical systems with guarantees on safety, robustness, and performance in the presence of disturbances and uncertainty. SOS programming relies on the notion of sum-of-squares polynomials. A polynomial is said to be a sum-of-squares if it can be written as a sum of squared polynomial terms. The set of SOS polynomials up to a certain degree $k$ is denoted $\Sigma_{k}[x]$. By definition,
\begin{align}
    P \in \Sigma_{k}[x] \iff  P(x) = \sum_{j=1}^{n}{g_{j}^{2}(x)} ,
\end{align}
for some polynomials $g_j(x)$. If a polynomial $P$ is SOS, its expression in terms of its associated $g_{i}(x)$ polynomials is called its SOS decomposition. Note that this decomposition is not unique in general.

The set $\Sigma_{k}[x]$ is a proper convex cone. As a consequence, general tools from convex optimization and conic programming can be directly applied to verify SOS conditions and decomposition \cite{Nocedal2006, Boyd2004}. Recently, the authors in \cite{Papp2018} have suggested the use of a primal-dual interior point method to solve SOS programs in a direct manner. Noting that the sum-of-squares cone is not symmetric and that currently no logarithmically homogeneous self-concordant barrier function is known for the primal cone, the authors leverage a non-symmetric conic solver. Ahmadi et. al. have also introduced the concepts of DSOS and SDSOS \cite{Ahmadi2019} that rely on linear and second-order-cone programming (LP \& SOCP) as approximations to sum-of-squares and semi-definite optimization. Up to now though, the widely-used alternative for dealing with SOS programs remains the exploitation of a simple, yet powerful link between SOS and Semi-Definite Programming (SDP) that can be summarized as,
\begin{align} \label{eq:sos-sdp}
    P \in \Sigma_{2k}[x] \iff P(x)=Z(x)^{T} Q Z(x), \ Q \in S_{k}^{+} ,
\end{align}
where $Z(x)$ is a vector of monomials in variable $x$ up to degree $k$ and $S_{k}^{+}$ is the set of positive semi-definite matrices of dimension $k$. Using this equivalence, certifying that a polynomial is SOS amounts to solving a positive-semi-definite feasibility program (i.e. finding a positive-semi-definite matrix, $Q$, satisfying \eqref{eq:sos-sdp}).

Sum-of-squares conditions are usually used as substitute conditions for non-negativity. While certifying non-negativity of a polynomial is NP-hard, sum-of-squares certificates can be obtained much more easily using semi-definite programming. Consequently, if a SOS certificate is found, non-negativity is obtained automatically. Note that the converse is not true in general. In addition, these techniques can be used even in the case of non-polynomial dynamical systems by performing a Taylor expansion on the dynamics function up to the desired degree. 

Note that we can extend the previous definitions and results to consider the general case of SOS certificates on semi-algebraic sets. For a more in depth treatment of SOS programming and its application to dynamics and control \cite{Singh} we refer the interested reader to the work of Parrilo \cite{Parrilo2000}.

\subsection{Lyapunov Stability Analysis}

Lyapunov analysis is concerned with showing convergence or stability properties of dynamical systems. Usually, the objective is to demonstrate the ability of a given controller to drive the system to a goal state, $x_g$, which can be assumed to be the origin of the system without loss of generality. The system is governed by a dynamical equation and a feedback controller,
\begin{align}
    \dot{x} &= g(x, u), \\
    u &= h(x). 
\end{align}
Where $x \in R^n$, is the state of the system and $u \in R^m$, is the control input. The composition of the controller with the system dynamics provides an autonomous closed-loop dynamics equation dictating the evolution of the system in time,
\begin{align}
    \dot{x} &= f(x).
\end{align}

A Lyapunov function, $V(x)$ is a positive function of the state of the system. This function is continuously differentiable over, $D$,  an open subset of $R^n$ containing the origin. Moreover, it respects the following global properties,
\begin{align}
    \label{eq:lyapunov_1}
    V(x) &> 0, 
    \quad \quad \quad \quad && \forall x \in D \setminus \{0\} \\
    \label{eq:lyapunov_2}
    V(0) &= 0,  \\
    \label{eq:lyapunov_3}
    \dot{V}(x) &= \frac{\partial V}{\partial x} f(x) < 0,
    \quad \quad \quad \quad && \forall x \in D \setminus \{0\} \\
    \label{eq:lyapunov_4}
    \dot{V}(0) &= 0. 
\end{align}
We encapsulate the Equations \ref{eq:lyapunov_1} - \ref{eq:lyapunov_4} with the following notations; $V \succ 0$ and $\dot{V} \prec  0$. 

$V$ is typically an energy-like function decreasing over time as the system evolves under its controlled dynamics. This Lyapunov function is used to demonstrate long-term behavior of the controlled system. In the literature, these techniques have been extended to non-autonomous dynamical systems as well.

\subsection{Region of Attraction}

A region-of-attraction, $G$, is a subset of $R^n$ for which the following property holds,
\begin{align}
    x(0) \in G \implies \lim_{t \rightarrow \infty} || x(t) - x_g || = 0.
\end{align}
Intuitively, this means that if the dynamical system starts inside $G$, it will converge to the goal state, $x_g$, as time $t$ goes to infinity. Sum-of-squares techniques are the basis for a large number of algorithms estimating regions of attractions for non-linear systems \cite{Meng2020}.

\subsection{Region of Finite-Time Invariance} 

It is possible to extend the notion of region of attraction in a timely manner to deal with regions of finite-time invariance, also known as invariant ``funnels''. As described in the work of Majumdar \cite{Majumdar2013, Majumdar2017}, a funnel can be seen as a time-varying generalization of a region of attraction. Loosely, an invariant funnel can be thought of as a tube around a reference trajectory within which a tracking controller is guaranteed to stabilize the system. Funnels have attracted widespread interest in the robotic motion planning community in recent years and have been used to solve a variety of challenging control problems.

\begin{figure}[htb]
	\centering\includegraphics[width=85mm]{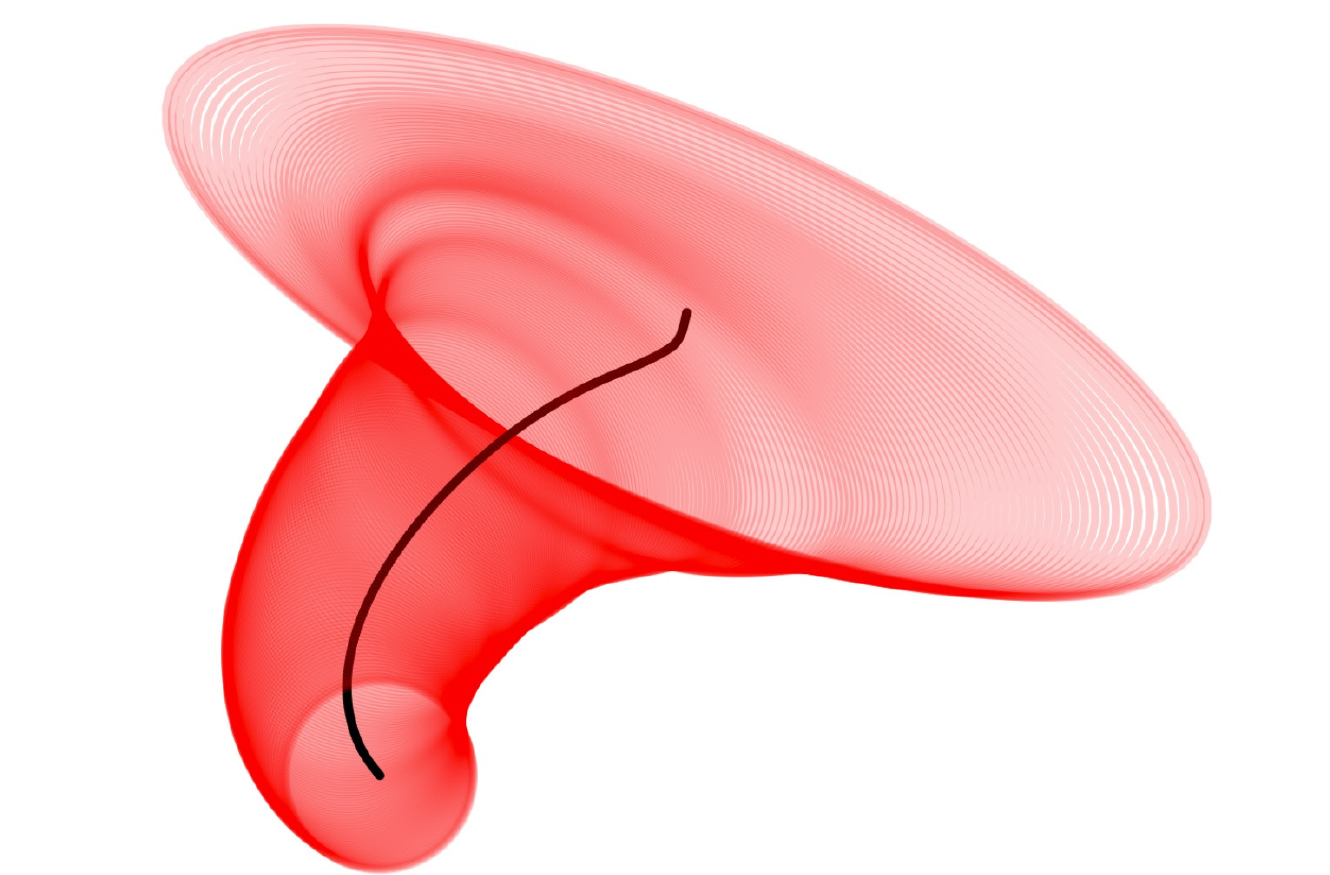}
	\caption{Three-dimensional invariant funnel computed around a nominal reference trajectory for a Dubins car.}
	\label{fig:dubins-funnel}
\end{figure}

Let's assume we are given a set of initial conditions $S_{1}$. We are interested in the set of goals that the system can evolve to at time $t \in [0, T]$. Mathematically we can define a funnel as a map $F$, 

\begin{align}
    F &: [0, T] \rightarrow P(R^{n}) \\
    F &: t \mapsto F(t) 
\end{align}

where the set $F(t)$ satisfies

\begin{align}
    \bar{x}(0) \in M_{1} \implies \bar{x}(t) \in F(t) \ \forall \ t  \in [0, T]
\end{align}

where $\bar{x}(t) = x(t) - x^{*}(t)$ is the difference between the current state and the reference state at time $t$. Just like SOS tools can be applied to compute approximations of regions of attraction for dynamical systems and specific controllers, they can also be used to compute inner and outer approximations of invariant funnels. 

In \cite{Tedrake2010} the authors compute approximate funnels around a nominal trajectory by defining $F(t)$ as,
\begin{align}
    F(t) &= \left\{ \bar{x}(t) | V(t, \bar{x}(t)) \leq \rho(t) \right\},
    \label{eq:funnel_framework}
\end{align}
where $\rho$ is a strictly positive function of time defining the boundary of the funnel and $V(t, \bar{x}(t))$ is a time-dependent Lyapunov function of the system under study. Figure \ref{fig:dubins-funnel} displays such a funnel computed around a nominal trajectory for a Dubins car system computed using the Julia package SystemVerification.jl developed by the authors.

This framework coupled with the SOS approach is a generic environment for computing invariant funnels in the literature \cite{Tobenkin2010}, and constitute the basis for the work developed in this paper. Other approaches exist that are related to the concept of invariant funnels, but rely on different tools. Tube Model Predictive Control (TMPC) allows for the design of robust controllers maintaining the system in a tube around the nominal trajectory \cite{Lopez2019, Li}. Reachability analysis is yet another framework following similar goals \cite{Bansal2017, Chen2016} with a different approach.

\vspace{-3mm}
\section{Sampling-based SOS Funnel}
\label{section:sampling-based_sos_funnel}

\subsection{Approach}

The authors in \cite{Cifuentes2017} develop the theory for using sampling on algebraic varieties to solve SOS programs while reducing computational effort. Building upon this theoretical foundation, Shen and Tedrake apply it for the computation of regions of attraction \cite{Shen}. The approach presented in this paper is inspired by their work and extends it to the computation of invariant funnels.

Following the traditional funnel computation described in \cite{Majumdar2017}, we use $({x}^{*}(t))_{t}$ to represent the reference trajectory followed by the system and $V(x, t)$ to represent a Lyapunov function of the system evaluated at state $x$ and time $t$. In the following, we consider a discrete-time version of the funnel. In this sense, we compute an approximation of the funnel by only focusing on ``slices'' at given time steps. We then reconnect the pieces together using interpolation techniques to obtain the complete funnel. In addition, in this work we focus on ellipsoidal-shaped funnels. In this configuration the Lyapunov functions are quadratic in $\bar{x}$ and we write,
\begin{align}
    V(\bar{x}, t_{k}) = \frac{1}{2}\bar{x}^{T}P(t_{k})\bar{x},
\end{align}
to refer to the Lyapunov function $V$ computed at time $t_{k}$ (corresponding to time step $k$) and evaluated at $\bar{x}$. 

To simplify notation we will write,
\begin{align}
    V_{k}(\bar{x}) &= V(\bar{x}, t_{k}) = V(x-x_{k}^{*}, t_{k}) \\
    x_{k} &= x(t_{k}) 
\end{align}
A slice of the funnel at a given time step $t_k$ is a mathematical set that can be written as,
\begin{align}
    F_{k} = F(t_{k}) = \{x | \frac{1}{2}(x-x^{*}_{k})^{T}P_{k}(x-x^{*}_{k})\leq \rho_{k} \}
    \label{eq:funnel_discrete_slice}
\end{align}
where $\rho_{k} = \rho(t_{k})$ is a scalar that concretely represents the limit of the funnel around the nominal trajectory at time step $k$. Equation (\ref{eq:funnel_discrete_slice})  is the discrete-time equivalent of Equation (\ref{eq:funnel_framework}). Note that, at this, stage $P_{k}$ and $x^{*}_{k}$ are known. Therefore the funnel is parameterized only by $\rho$.

We consider computing funnels in a forward manner. Using our discrete definition, we can compute the volume of the funnel with,
\begin{align}
   S = \sum_{k=1}^{N} \rho_k^{n} \det(P_k)^{-1},
\end{align}
which basically corresponds to the sum of the volumes of the ellipsoidal slices in time. The quantity $S$ represents the objective of our optimization problem. 

Let's now focus on the definition of the constraints of the optimization problem that we are trying to solve. We can distinguish two different types of constraints: The first one corresponds to the constraint at the initial time step that is needed to start the computation of the funnel. We require that the initial region of the state-space in which the system can be found be included in the first slice of the funnel parameterized by the positive semi-definite matrix $P_{1}$. By considering an ellipsoidal initial region $S_{1}$ parameterized by the matrix $M_{1}$, a simple derivation shows that this constraint can be written,
\begin{align}
    M_1 - \frac{P_1}{\rho_1} &\succeq 0,
    \label{eq:initial_constraint}
\end{align}
which is a simple semi-definite constraint allowing us to find $\rho_{1}$. 

The second type of constraint is the stage constraint that allows us to compute the value of $\rho_{k+1}$ given the value of $\rho_{k}$. On the boundary of the funnel at $t_k$, we require $\dot{\rho}(t_k) > \dot{V}(t_k, \bar{x})$. We denote the boundary of the funnel by $B_k$ where
\begin{align}
   B_k &= \left\{ \bar{x}| V(t_k, \bar{x}) = \rho_k \right\}.
   \label{eq:stage_constraint_1}
\end{align}
This condition can be expanded to obtain a relation between $\rho_{k+1}$ and $\rho_{k}$ such that,
\begin{align}
\frac{\rho_{k+1}}{t_{k+1} - t_k} &>
        \frac{1}{2} \frac{\bar{x}^T P_{k+1} \bar{x}}{t_{k+1} - t_k} + \ldots \nonumber\\
        &{\left[f_k(\bar{x} + x_k^*) - f_k(x_k^*) \right]}^T
        P_k \bar{x},
        \label{eq:stage_constraint_2}
\end{align}
where $f_k$ is the closed-loop dynamics function of the system evaluated with control inputs from time step $k$ leveraging the linear controller gain $K_{k}$ obtained with iLQR. The derivation can be found in Appendix A.

We rewrite Equation (\ref{eq:stage_constraint_2}) in the following way for simplicity: 
\begin{align}
    D_k(\bar{x}; \rho_{k+1}) &> 0.
        \quad \quad \forall \: \bar{x} \in B_k
    \label{eq:stage_constraint_3}
\end{align}
This notation shows that at each time step $k \geq 1$, we can determine $\rho_{k+1}$ by using $P_{k}, P_{k+1}$ and $f_k$. It also shows that at each time step, the condition must hold for all values of $\bar{x}$ in $B_k$.

Assuming that the quantity $D_{k}$ in Equation (\ref{eq:stage_constraint_3}) can be approximated by a polynomial function (using a Taylor expansion to the appropriate order, for instance), and that the set $B_k$ is an algebraic variety (which is always the case given our assumptions), we can relax the non-negativity constraint in Equation (\ref{eq:stage_constraint_3}) to a sum-of-squares constraint. Traditionally, the so-called ``multiplier approach'' is used to incorporate the constraint on the set in the inequality and solve the problem in a global manner. 

This is where our approach diverges: We use the quotient-ring sampling strategy \cite{Shen, Cifuentes2017} to rewrite the constraint in Equation (\ref{eq:stage_constraint_3}). Indeed, following the theoretical results provided in the references, we can rewrite the SOS relaxed constraint in the following \emph{equivalent} way (provided that the proper assumptions hold),
\begin{align}
    {({\bar{x}_k^{(i)\:T}}  \bar{x}^{(i)}_k)}^d  
    D_k(\bar{x}^{(i)}_k)
    = 
    {\tilde{n}(\bar{x}^{(i)}_k)}^T Q_k \tilde{n}(\bar{x}^{(i)}_k),
\label{eq:stage_constraint_final}
\end{align}
where $Q_k$ is a new positive semi-definite variable added to the problem, $d$ is an arbitrary integer exponent and $(\bar{x}^{(i)}_k)$ is a family of $M_{k}$ well chosen samples drawn from $B_{k}$ and $\tilde{n}$ is a vector containing a polynomial basis. We end up with a semi-definite feasibility constraint (on the variable $Q_k$) to be verified on a specific set of samples.

To summarize, the optimization problem that we solve can be written,
\begin{mini}|l|
    {\rho_1, \ldots, \rho_N, Q_1, \ldots, Q_{N-1}}
    {\sum_{k=1}^{N} \rho_k^{n} \det(P_k)^{-1}}{}{}
    \addConstraint{M_1 - \frac{P_1}{\rho_1} \succeq 0}
    \addConstraint{\bar{x}^{(i)}_k \in B_k, \quad \forall k \in \llbracket 1,N-1 \rrbracket}
    \addConstraint{ {({\bar{x}^{(i)\:T}_k}  \bar{x}^{(i)}_k)}^d  D_k(\bar{x}^{(i)}_k ; \rho_{k+1})}
    \addConstraint{= {\tilde{n}(\bar{x}^{(i)}_k)}^T Q_k \tilde{n}(\bar{x}^{(i)}_k) }
    \addConstraint{\quad \quad \forall \bar{x}^{(i)}_k, \ \forall k \in \llbracket 1,N-1 \rrbracket},
    \label{opt:global_problem}
\end{mini}
where $N$ is the total number of steps. 

One important difference compared to the multiplier approach is that the algebraic variety, $B_{k}$, that needs to be sampled depends on the value of $\rho_{k}$. Therefore, we need to know this value in order to be able to sample the 
$\bar{x}^{(i)}_k$ values and obtain a constraint on $\rho_{k+1}$. This prevents us from a global approach and brings us to consider a sequential "greedy" strategy at each time step to obtain the successive values of $\rho_{k}$ with $\rho_{1}$ given by the initial constraint in Equation (\ref{eq:initial_constraint}).

\subsection{Sampling Strategy}


An important step in the approach described above is the sampling of the points on the variety $B_{k}$ at each time step. Different approaches are described for sampling the points $\bar{x}_k^{(i)}$ on the variety $B_k$ and we refer the reader to the work of Cifuentes \cite{Cifuentes2017} for a detailed and rigorous treatment of the topic.

In this paper, we note that we can choose a random point $\tilde{x}^{(i)}$ in $R^n$, then rescale it to obtain our sample $\bar{x}^{(i)}$ , which belongs to the variety automatically:
\begin{align}
    \label{eq:sampling_x_bar}
    \bar{x}^{(i)} &= \frac{\tilde{x}^{(i)}}{\sqrt{\frac{1}{2} (\tilde{x}^{(i)})^T \frac{P_k}{\rho_k} \tilde{x}^{(i)}}} ,\\
    1 &= \frac{1}{2} (\tilde{x}^{(i)})^T \frac{P_k}{\rho_k} \tilde{x}^{(i)} \implies \bar{x}^{(i)} \in B_k.
\end{align}

Interestingly, the minimum number of sample points necessary to guarantee equivalence between inequalities (\ref{eq:stage_constraint_3}) and (\ref{eq:stage_constraint_final}) can be computed. We implement the technique described in \cite{Cifuentes2017} to find that minimum number of sample points and apply it for real vectors in our case. This indicator gives us a threshold on the number of samples to gather on the variety.

When analytical sampling methods are not applicable, we notice that the algebraic varieties $(B_{k})_{k}$ are defined by polynomial equalities only. This amounts to finding a subset of the roots of multivariate polynomials. Root-finding techniques like Newton's method can therefore be applied to sample these sets.

\subsection{Disturbances}

In the formulation described above, we have assumed that an initial state-space region $S_1$ was given to represent the uncertainty on the state of the system. In a similar way, we would like to be able to handle disturbances and uncertainty on parameters of the dynamical model. Following the same robust approach, we define a vector $w$ to represent the uncertain parameters in the model. We also assume that these uncertain parameters live in an algebraic set given by a quadratic function and parameterized by a matrix $U$:
\begin{align}
    W = \left\{ w \: | \: \frac{1}{2} w^T U w \leq 1 \right\}.
\end{align} 

To extend the computation of the invariant funnel under the disturbance $w$, we refer again to the work of Majumdar \cite{Majumdar2017}. The invariant funnel condition can be expressed in continuous time as follows, 

\begin{align}
    \label{eq:continuous_dist_funnel}
    \dot{\rho}(t) &> \dot{V}(t,\bar{x},w) 
    \quad \quad \forall \: \bar{x} \in B(t) \quad \forall t \in [0, T] \quad \forall w \in W, \\
    V(t) &= \left\{ \bar{x}| V(t, \bar{x}) = \rho(t) \right\}.
\end{align}

Verifying the condition in Equation (\ref{eq:continuous_dist_funnel}) on the discretized time steps $t_k$'s, would require verifying the inequality for any $w \in W$ at each step. Again, instead of making use of the multiplier approach, we can adapt the sampling technique to draw samples from the relevant variety.

Intuitively, we want to make sure that, by propagating the dynamics under any feasible disturbance for one time step, we end up below the threshold $\rho(t_{k+1})$. Since the dynamics $f_k(x,w)$ is a polynomial function of all its entries, it is continuous and differentiable with respect to $w$ for any $x$. The same is true for $\dot{V}_k$. Therefore, using Fermat's theorem, we know that the extremal values obtained for $\dot{V}_k$ are located either on the boundary of $W$, denoted $\partial W$, or on points $(\bar{x},w)$ such that $\frac{\partial \dot{V}_k}{\partial w} (\bar{x},w) = 0 $. More formally, the relevant disturbances are located in the union of two sets,
\begin{align}
    w^* &\in \text{argmax}_{w} \dot{V}_k(\bar{x}, w) \implies  w \in \partial W \cup O(\bar{x}),
\end{align}
where we define 
\begin{align}
    \partial W &= \left\{ w \: | \: g(w) = 0 \right\}, \\
    O(\bar{x}) &= \left\{ w \: | \: \frac{\partial \dot{V}}{\partial w}(\bar{x}, w) = 0 \right\}.
\end{align}
We remark that both $\partial W$ and $O(\bar{x})$ are algebraic varieties and, therefore, the sampling procedure for SOS constraints can be applied to these sets as well.

We now describe the sampling strategy taking into account uncertainty parameters and disturbances. First, we sample $\bar{x}$  using the same sampling strategy as in the disturbance-free case. Given this sampled $\bar{x}$, we then chose to sample $w$ either from $\partial W$ or $O(\bar{x})$. We can proceed as follows for the first set: 
\begin{align}
    w = \frac{\tilde{w}}{\sqrt{\frac{1}{2} \tilde{w}^T W \tilde{w}}} \implies w \in \partial W, \quad \forall \: \tilde{w} \in R^p .
    \label{eq:sampling_w_1}
\end{align}
For the second set, we can derive (see Appendix B):
\begin{align}
    \frac{\partial \dot{V}}{\partial w}(\bar{x}, w) = \frac{\partial}{\partial w} 
        {\left[ f_k(\bar{x}+x_k^*, w) \right]}^T P_k \bar{x}.
    \label{eq:sampling_w_2}
\end{align}

We have access to the polynomial expression of the feedback dynamics $f_k$ and to its derivative with respect to $w$. Given $\bar{x}$, we can apply a root-finding algorithm, like Newton's method, to find a suitable $w \in O(\bar{x})$. Similarly to what has been described in the Approach Section, we simply need to verify the SOS constraints on a small number of samples, allowing us to drastically reduce the computational burden compared to traditional SOS approaches.

The full procedure is summarized the procedure in Algorithm \ref{alg:sampling_sos_funnel}, where \textbf{SDP} refers to solving a semi-definite program defined as, 
\begin{mini}|l|
    {\rho_{k+1}, Q_{k}}
    {f(\rho_{k+1})}{}{}
    \addConstraint{\bar{x}^{(i)}_k \in B_k, \quad \quad \forall i \in \llbracket 1,M_{k} \rrbracket}
    \addConstraint{ {({\bar{x}^{(i)\:T}_k}  \bar{x}^{(i)}_k)}^d  D_k(\bar{x}^{(i)}_k; \rho_{k+1})}
    \addConstraint{= {\tilde{n}(\bar{x}^{(i)}_k)}^T Q_k \tilde{n}(\bar{x}^{(i)}_k) }
    \addConstraint{\quad \quad \forall \bar{x}^{(i)}_k},
    \label{opt:reduced_problem}
\end{mini}
where $Q_{k}$ is a positive semi-definite matrix and $f$ a simple positive and increasing function. Note that the minimization is related to the forward funnel computation. A backward approach would feature maximization.

\begin{algorithm}
    \SetKwInOut{Input}{Input}
    \SetKwInOut{Output}{Output}

    \Input{$(t_{k}, P_{k}, x_k^{*}, u_k^{*})_k, M_{1}, U$}
    \textbf{Procedure} \\

    Obtain minimum $\rho_1$ satisfying (\ref{eq:initial_constraint}); \\
    \For{$k=1$ \ \text{to} \ $N-1$}
      {
        $\quad (\tilde{x}_k^{(i)})_i \gets \textbf{SAMPLE STATE}(P_{k}, \rho_{k}, x_k^{*})$\;
        $\quad (\tilde{w}_k^{(j)})_j \gets \textbf{SAMPLE DIST}(P_{k}, \rho_{k}, x_k^{*}, (\tilde{x}_k^{(i)})_i)$\;
        $\quad \rho_{k+1} \gets 
        \textbf{SDP}(\rho_{k}, P_{k}, P_{k+1}, x_k^{*}, (\tilde{x}_k^{(i)})_i, (\tilde{w}_k^{(j)})_j)$\;
      }
      {
        \textbf{return} \ $(\rho_{k})_{k}$ \\
      }
      \textbf{End Procedure}
    \caption{Sampling SOS Funnel Computation}
    \label{alg:sampling_sos_funnel}
\end{algorithm}

\vspace{-3mm}
\section{Simulation Results}
\label{section:simulation_results}

In this section, we present the results of the sampling funnel approach applied to the entry guidance problem. There are several ways to derive the required inputs (i.e. nominal trajectory, $x_{k}^{*}$ and time-dependent Lyapunov function $V_{k}$) to the process described in Algorithm \ref{alg:sampling_sos_funnel}. Our simulations use ALTRO \cite{Howell}, a fast trajectory optimization solver, to find an optimal reference trajectory and the optimal control sequence to reach a specific target. A time-varying linear-quadratic regulator (TVLQR) is then used to obtain a quadratic value function approximation. The positive semi-definite matrix parameterizing this value function is then used in place of the Lyapunov functions for the funnel computation. The trajectory optimization ensures that thermal constraints are satisfied \cite{Bose2013}.

We consider a three-degree-of-freedom dynamical model for the entry. We recall Vinh's model below \cite{Vinh1980, ManriqueJoelBenito2010}. This representation uses the position vector of the center of mass of the vehicle in spherical coordinates and flight path and heading angles for the direction of the velocity vector,
\begin{align}
& \dot{r} = V\sin{\gamma} \nonumber\\
& \dot{\theta} = \frac{V}{r} \frac{\cos{\gamma}\cos{\psi}}{\cos{\phi}} \nonumber \\
& \dot{\phi} = \frac{V}{r}\cos{\gamma}\sin{\psi} \nonumber \\
& \dot{V} = -D-g\sin{\gamma} \\
& \dot{\gamma} = \frac{1}{V}[L\cos{\sigma}-(g-\frac{V^{2}}{r})\cos{\gamma}]+C_{\gamma} \nonumber \\
& \dot{\psi} = -\frac{1}{V\cos{\gamma}}(L\sin{\sigma}+\frac{V^{2}}{r}\cos^{2}{\gamma}\cos{\psi}\tan{\phi})+C_{\psi} \nonumber, 
\end{align}
where $C_{\gamma}$, $C_{\psi}$ are the Coriolis terms and $L$ and $D$ are lift and drag forces, respectively.

Using this representation, the state space has dimension 6 and the state vector is written $X=[r, \ \theta, \ \phi, \ V, \ \gamma,\ \psi]$ where $r, \theta, \phi$ is the position vector expressed in spherical coordinates with origin at the center of the planet, $V$ is the magnitude of the velocity vector, while $\gamma$ and $\psi$ are respectively the flight path and heading angle giving the direction of the velocity vector. Vehicle parameters and a nominal trajectory approximately matching the Mars Science Laboratory (MSL) mission were used for this study. The simulation was stopped 10 km above the surface corresponding to parachute deployment. 

Uncertainties are taken into account in the state of the spacecraft as well as in the atmospheric parameters. Relevant additive external disturbances are present in all components of the dynamical model modeling position uncertainty and winds. Finally, uncertainty is considered in the atmospheric density parameters. MarsGRAM 2010 is used to infer proper bounds on the parameters of a bilinear exponential density model. 

An open-source Julia package was developed by the authors to perform these funnel computations called SystemVerification.jl. We have made our code available at \href{https://github.com/RoboticExplorationLab/SystemVerification.jl}{\texttt{github.com/RoboticExplorationLab/SystemVe \\ rification.jl}}. Algorithm \ref{alg:sampling_sos_funnel} along with a multiplier-based funnel solver, a Monte Carlo funnel solver and a linear dynamics ellipsoidal funnel solver \cite{Manchester} are implemented in the package. The optimization solver Mosek \cite{mosek} is used to solve SDPs where needed. At all time steps, the value of the parameter $d$ in Equation ($\ref{eq:stage_constraint_3}$) is set to 1. Consequently, $Q_{k}$ is a 45-dimensional square matrix at each time step. When drawing samples on a variety is not straightforward and requires finding roots of a multivariate polynomial (see Section \ref{section:sampling-based_sos_funnel}), a custom Newton solver is used. For good numerical conditioning, we work with the Hermite polynomial basis for $\tilde{n}$.

\begin{figure}
\centering
\includegraphics[width=8.50cm, height=6.00cm]{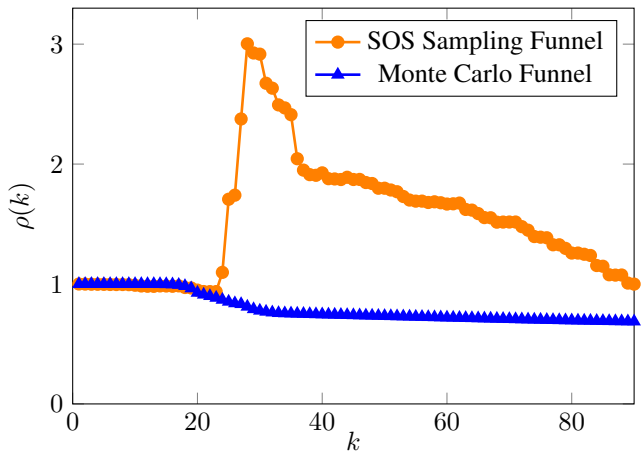}\\
\caption{Funnel boundary $\rho$ vs time step for both SOS-sampling and Monte Carlo approaches on the entry guidance problem.
\vspace{-6mm}
}
\label{fig:funnel-rho}
\end{figure}

\begin{figure}
\centering
\includegraphics[width=8.50cm, height=6.00cm]{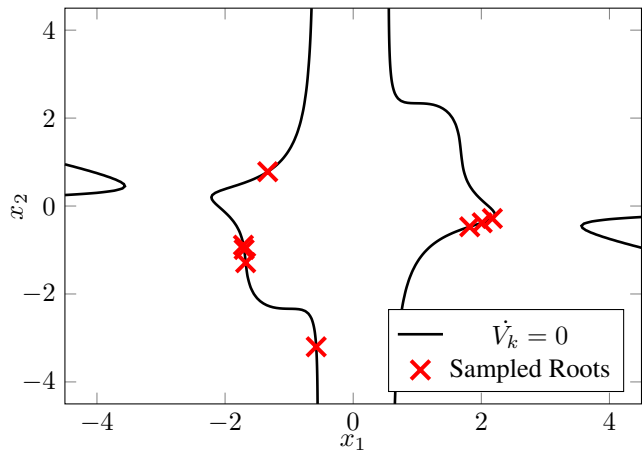}\\
\caption{Sampled roots location for $k = 7$ on the variety $B_{k}$ covering the 0-sublevel set of the time derivative of the Lyapunov function $V_{k}$.
\vspace{-6mm}
}
\label{fig:roots-location}
\end{figure}

Figure \ref{fig:funnel-rho} presents the value of $\rho$ with respect to time for both the Sampling-SOS approach and a traditional Monte Carlo approach. $10^{4}$ trajectories were generated and propagated for the Monte Carlo run. Figure \ref{fig:roots-location} shows the location of the sampled points (roots of multivariate polynomial) for a 2D slice of the state space along with the 0 sub-level set of the time derivative of the Lyapunov function for a specific time step.

\vspace{-3mm}
\section{Discussion}
\label{section:discussion}

Figure \ref{fig:funnel-rho} allows us to validate the approach and to compare it to traditional techniques. A first peak at the beginning of the trajectory is observed corresponding to the vehicle entering the atmosphere, and to the corresponding large deceleration that increases the uncertainty and the volume of the funnel.
The invariant funnel also expands towards the end of the trajectory while approaching the ground, where high velocity changes occur. In order to analyze these results properly, it is worth mentioning the major difference that lies in the meaning of these envelopes and that explains the discrepancies observed between our method and Monte Carlo. While traditional approaches provides a result under specific probabilistic assumptions (Gaussian distributions implicitly assumed most of the time), SOS techniques do not assume any particular distribution and instead capture strict bounds on sets. Figure \ref{fig:funnel-rho} clearly shows that all Monte Carlo sample trajectories are included in the region drawn using our conservative approach. Monte Carlo solely provides statistical information with no guarantees. 


The robustness and non-probabilistic properties also hold in the case of traditional multiplier SOS approaches.
However, a multiplier-based SOS approach is prohibitively expensive in the entry vehicle case. Problems with more than 6 or 7 dimensions, including states, controls, and disturbances, are usually intractable with these techniques unless some special separable structure exists in the problem. In comparison, our sampling-based approach verifies equivalent SOS constraints only on a small number of samples. The computation time for our entry funnel example is approximately one minute on a standard laptop computer.

Note that we have assumed quadratic Lyapunov functions in this analysis and have obtained ellipsoidal regions of finite-time invariance in consequence. The theoretical validity of the development presented in Section \ref{section:sampling-based_sos_funnel} can be extended to more complex algebraic sets. We have also focused on funnel computation in a forward manner. Backward funnel computation is also possible. A backward version of the sampling-based SOS funnel solver is implemented in SystemVerification.jl.

\vspace{-3mm}
\section{Conclusions}

We have presented a set-based robust approach for uncertainty propagation in entry guidance applications. Our formulation relies on a recent sampling-based sum-of-squares method that is extended to the computation of invariant funnels. The results of this new technique are compared to a classic Monte Carlo approach. The new method does not rely on probabilistic assumptions and provides hard safety and performance guarantees during entry, despite external disturbances and uncertain model parameters, while drastically reducing computational effort. The software developed by the authors has been released in the open-source SystemVerification.jl package for invariant funnel computation.

Directions for future work include extending this analysis to a six-degree-of-freedom entry vehicle model taking into account attitude dynamics. There are also interesting possibilities for combining our invariant funnels with online MPC strategies for improved performance while still maintaining safety guarantees.
Finally, given the power and performance of sampling-based SOS techniques, further developments should also focus on improving computational performance for potential use in online contexts. 

\vspace{-3mm}
\appendices{}

\section{Stage constraint derivation}        

We provide the derivation for the equations used in the sampling version of the funnel:
\vspace{-2mm}
\begin{align*}
    \dot{\rho}(t_k) &> \dot{V}(t_k, \bar{x}), 
        \quad \quad \forall \: \bar{x} \in B_k \\
    \implies \frac{\rho_{k+1} - \rho_k}{t_{k+1} - t_k} &>
        \frac{V_{k+1}(\bar{x}) - V_k(\bar{x})}{t_{k+1} - t_k} + 
        {\frac{\partial \bar{x}}{\partial t}\Bigr|}^T_{t=t_k}
        \frac{\partial V_k}{\partial x}(\bar{x}), \\
    \implies \frac{\rho_{k+1}}{t_{k+1} - t_k} &>
        \frac{V_{k+1}(\bar{x})}{t_{k+1} - t_k} + 
        {\frac{\partial (x - x_k^*)}{\partial t}\Bigr|}^T_{t=t_k}
        \frac{\partial V_k}{\partial x}(\bar{x}), \\
    \implies \frac{\rho_{k+1}}{t_{k+1} - t_k} &>
        \frac{1}{2} \frac{\bar{x}^T P_{k+1} \bar{x}}{t_{k+1} - t_k} +  \ldots \\
        &{\left[f(x, K_k x) - f(x_k^*, K_k x_k^*) \right]}^T
        P_k \bar{x}, \\
    \implies \frac{\rho_{k+1}}{t_{k+1} - t_k} &>
        \frac{1}{2} \frac{\bar{x}^T P_{k+1} \bar{x}}{t_{k+1} - t_k} + \ldots \\
        &{\left[f_k(\bar{x} + x_k^*) - f_k(x_k^*) \right]}^T
        P_k \bar{x}, \\
    \implies  D_k(\bar{x}; \rho_{k+1}) &> 0
        \quad \quad \forall \: \bar{x} \in B_k ,
\end{align*}
where $f_k$ is the closed loop dynamics of the system using the linear control $K_k $. Note that the inequalities must hold $\forall \: \bar{x} \in B_k$.

\vspace{-3mm}
\section{Disturbances sampling derivation}        

We present the derivation for the disturbance sampling strategy on the algebraic variety $O$:
\vspace{-2mm}

\begin{align*}
    \frac{\partial \dot{V}}{\partial w}(\bar{x}, w) &= \frac{\partial}{\partial w} \left[ 
        \frac{\partial V}{\partial t}(t_k, \bar{x}) +  
        {\frac{\partial \bar{x}}{\partial t}(t_k, \bar{x}, w)}^T \frac{\partial V}{\partial \bar{x}}(t_k, \bar{x})
        \right], \\
    &= \frac{\partial}{\partial w} \left[ 
        {\frac{\partial \bar{x}}{\partial t}(t_k, \bar{x}, w)}^T \frac{\partial V}{\partial \bar{x}}(t_k, \bar{x})
        \right], \\
    &= \frac{\partial}{\partial w} \left[ 
        {\frac{\partial \bar{x}}{\partial t}(t_k, \bar{x}, w)}^T P_k \bar{x}
        \right], \\
    &= \frac{\partial}{\partial w} 
        {\left[ \frac{\partial \bar{x}}{\partial t}(t_k, \bar{x}, w) \right]}^T P_k \bar{x}, \\
    &= \frac{\partial}{\partial w} 
        {\left[ \frac{\partial x(t)}{\partial t}(t_k, x, w) - \frac{\partial x^*(t)}{\partial t}(t_k, x, 0) \right]}^T P_k \bar{x}, \\
    &= \frac{\partial}{\partial w} 
        {\left[ f_k(x, w) - f_k(x_k^*, 0) \right]}^T P_k \bar{x}, \\
    &= \frac{\partial}{\partial w} 
        {\left[ f_k(\bar{x}+x_k^*, w) - f_k(x_k^*, 0) \right]}^T P_k \bar{x}, \\
    \label{eq:sampling_w}
    &= \frac{\partial}{\partial w} 
        {\left[ f_k(\bar{x}+x_k^*, w) \right]}^T P_k \bar{x}.
\end{align*}

\acknowledgments
This work was supported by an Early Career Faculty grant from NASA’s Space Technology Research Grants Program.

\newcommand{\newblock}{}
\bibliographystyle{IEEEtran}
\bibliography{reference}

\begin{thebibliography}{10}
\providecommand{\url}[1]{#1}
\csname url@samestyle\endcsname
\providecommand{\newblock}{\relax}
\providecommand{\bibinfo}[2]{#2}
\providecommand{\BIBentrySTDinterwordspacing}{\spaceskip=0pt\relax}
\providecommand{\BIBentryALTinterwordstretchfactor}{4}
\providecommand{\BIBentryALTinterwordspacing}{\spaceskip=\fontdimen2\font plus
\BIBentryALTinterwordstretchfactor\fontdimen3\font minus
  \fontdimen4\font\relax}
\providecommand{\BIBforeignlanguage}[2]{{%
\expandafter\ifx\csname l@#1\endcsname\relax
\typeout{** WARNING: IEEEtran.bst: No hyphenation pattern has been}%
\typeout{** loaded for the language `#1'. Using the pattern for}%
\typeout{** the default language instead.}%
\else
\language=\csname l@#1\endcsname
\fi
#2}}
\providecommand{\BIBdecl}{\relax}
\BIBdecl

\bibitem{Li2014}
S.~Li and X.~Jiang, ``\BIBforeignlanguage{en}{Review and prospect of guidance
  and control for {{Mars}} atmospheric entry},''
  \emph{\BIBforeignlanguage{en}{Progress in Aerospace Sciences}}, vol.~69, pp.
  40--57, Aug. 2014.

\bibitem{Braun2006}
R.~Braun and R.~Manning, ``\BIBforeignlanguage{en}{Mars {{Exploration Entry}},
  {{Descent}} and {{Landing Challenges}}},'' in
  \emph{\BIBforeignlanguage{en}{2006 {{IEEE Aerospace Conference}}}}.\hskip 1em
  plus 0.5em minus 0.4em\relax {Big Sky, MT, USA}: {IEEE}, 2006, pp. 1--18.

\bibitem{tauber1991stagnation}
M.~E. Tauber and K.~Sutton, ``Stagnation-point radiative heating relations for
  {{Earth}} and {{Mars}} entries,'' \emph{Journal of Spacecraft and Rockets},
  vol.~28, no.~1, pp. 40--42, 1991.

\bibitem{Bose2013}
D.~Bose, J.~L. Brown, D.~K. Prabhu, P.~Gnoffo, C.~O. Johnston, and B.~Hollis,
  ``\BIBforeignlanguage{en}{Uncertainty {{Assessment}} of {{Hypersonic
  Aerothermodynamics Prediction Capability}}},''
  \emph{\BIBforeignlanguage{en}{Journal of Spacecraft and Rockets}}, vol.~50,
  no.~1, pp. 12--18, Jan. 2013.

\bibitem{zang2010overview}
T.~Zang, A.~Dwyer~Cianciolo, D.~Kinney, A.~Howard, G.~Chen, M.~Ivanov,
  R.~Sostaric, and C.~Westhelle, ``Overview of the nasa entry, descent and
  landing systems analysis stud,'' in \emph{{{AIAA SPACE}} 2010 Conference \&
  Exposition}, 2010, p. 8649.

\bibitem{Roenneke2001}
A.~Roenneke, ``\BIBforeignlanguage{en}{Adaptive on-board guidance for entry
  vehicles},'' in \emph{\BIBforeignlanguage{en}{{{AIAA Guidance}},
  {{Navigation}}, and {{Control Conference}} and {{Exhibit}}}}.\hskip 1em plus
  0.5em minus 0.4em\relax {Montreal,Canada}: {American Institute of Aeronautics
  and Astronautics}, Aug. 2001.

\bibitem{Li2011}
S.~Li and Y.~Peng, ``\BIBforeignlanguage{en}{Mars entry trajectory optimization
  using {{DOC}} and {{DCNLP}}},'' \emph{\BIBforeignlanguage{en}{Advances in
  Space Research}}, vol.~47, no.~3, pp. 440--452, Feb. 2011.

\bibitem{Saraf2004}
A.~Saraf, J.~Leavitt, M.~Ferch, and K.~Mease, ``\BIBforeignlanguage{en}{Landing
  {{Footprint Computation}} for {{Entry Vehicles}}},'' in
  \emph{\BIBforeignlanguage{en}{{{AIAA Guidance}}, {{Navigation}}, and
  {{Control Conference}} and {{Exhibit}}}}.\hskip 1em plus 0.5em minus
  0.4em\relax {Providence, Rhode Island}: {American Institute of Aeronautics
  and Astronautics}, Aug. 2004.

\bibitem{benito2010reachable}
J.~Benito and K.~D. Mease, ``Reachable and controllable sets for planetary
  entry and landing,'' \emph{Journal of guidance, control, and dynamics},
  vol.~33, no.~3, pp. 641--654, 2010.

\bibitem{Rahimi2013}
A.~Rahimi, K.~Dev~Kumar, and H.~Alighanbari, ``\BIBforeignlanguage{en}{Particle
  {{Swarm Optimization Applied}} to {{Spacecraft Reentry Trajectory}}},''
  \emph{\BIBforeignlanguage{en}{Journal of Guidance, Control, and Dynamics}},
  vol.~36, no.~1, pp. 307--310, Jan. 2013.

\bibitem{Steinfeldt2010}
B.~Steinfeldt and P.~Tsiotras, ``\BIBforeignlanguage{en}{A
  {{State}}-{{Dependent Riccati Equation Approach}} to {{Atmospheric Entry
  Guidance}}},'' in \emph{\BIBforeignlanguage{en}{{{AIAA Guidance}},
  {{Navigation}}, and {{Control Conference}}}}.\hskip 1em plus 0.5em minus
  0.4em\relax {Toronto, Ontario, Canada}: {American Institute of Aeronautics
  and Astronautics}, Aug. 2010.

\bibitem{Li2015}
S.~Li and X.~Jiang, ``\BIBforeignlanguage{en}{{{RBF}} neural network based
  second-order sliding mode guidance for {{Mars}} entry under uncertainties},''
  \emph{\BIBforeignlanguage{en}{Aerospace Science and Technology}}, vol.~43,
  pp. 226--235, Jun. 2015.

\bibitem{luo2017review}
Y.-z. Luo and Z.~Yang, ``A review of uncertainty propagation in orbital
  mechanics,'' \emph{Progress in Aerospace Sciences}, vol.~89, pp. 23--39,
  2017.

\bibitem{lockwood2001entry}
M.~K. Lockwood, R.~W. Powell, C.~A. Graves, and G.~L. Carman, ``Entry system
  design considerations for {{Mars}} landers,'' \emph{AAS Guidance and Control
  Conference}, 2001.

\bibitem{jones2013nonlinear}
B.~A. Jones, A.~Doostan, and G.~H. Born, ``Nonlinear propagation of orbit
  uncertainty using non-intrusive polynomial chaos,'' \emph{Journal of
  Guidance, Control, and Dynamics}, vol.~36, no.~2, pp. 430--444, 2013.

\bibitem{Prabhakar2010}
A.~Prabhakar, J.~Fisher, and R.~Bhattacharya,
  ``\BIBforeignlanguage{en}{Polynomial {{Chaos}}-{{Based Analysis}} of
  {{Probabilistic Uncertainty}} in {{Hypersonic Flight Dynamics}}},''
  \emph{\BIBforeignlanguage{en}{Journal of Guidance, Control, and Dynamics}},
  vol.~33, no.~1, pp. 222--234, Jan. 2010.

\bibitem{Derollez}
R.~Derollez and Z.~Manchester, ``\BIBforeignlanguage{en}{Sample-based {{Robust
  Uncertainty Propagation}} for {{Entry Vehicles}}},''
  \emph{\BIBforeignlanguage{en}{AAS/AIAA Astrodynamics Specialist Conference}},
  2020.

\bibitem{jin2018development}
K.~Jin, D.~Geller, and J.~Luo, ``Development and validation of linear
  covariance analysis tool for atmospheric entry,'' \emph{Journal of Spacecraft
  and Rockets}, vol.~56, no.~3, pp. 854--864, 2018.

\bibitem{Woffinden2019}
D.~Woffinden, S.~Robinson, J.~Williams, and Z.~R. Putnam,
  ``\BIBforeignlanguage{en}{Linear {{Covariance Analysis Techniques}} to
  {{Generate Navigation}} and {{Sensor Requirements}} for the {{Safe}} and
  {{Precise Landing Integrated Capabilities Evolution}} ({{SPLICE}})
  {{Project}}},'' in \emph{\BIBforeignlanguage{en}{{{AIAA Scitech}} 2019
  {{Forum}}}}.\hskip 1em plus 0.5em minus 0.4em\relax {San Diego, California}:
  {American Institute of Aeronautics and Astronautics}, Jan. 2019.

\bibitem{Tedrake2010}
R.~Tedrake, I.~R. Manchester, M.~Tobenkin, and J.~W. Roberts,
  ``\BIBforeignlanguage{en}{{{LQR}}-trees: {{Feedback Motion Planning}} via
  {{Sums}}-of-{{Squares Verification}}},'' \emph{\BIBforeignlanguage{en}{The
  International Journal of Robotics Research}}, vol.~29, no.~8, pp. 1038--1052,
  2010.

\bibitem{Majumdar2017}
A.~Majumdar and R.~Tedrake, ``\BIBforeignlanguage{en}{Funnel {{Libraries}} for
  {{Real}}-{{Time Robust Feedback Motion Planning}}},''
  \emph{\BIBforeignlanguage{en}{arXiv:1601.04037 [cs, math]}}, 2017.

\bibitem{Tobenkin2010}
M.~M. Tobenkin, I.~R. Manchester, and R.~Tedrake,
  ``\BIBforeignlanguage{en}{Invariant {{Funnels}} around {{Trajectories}} using
  {{Sum}}-of-{{Squares Programming}}},''
  \emph{\BIBforeignlanguage{en}{arXiv:1010.3013 [math]}}, 2010.

\bibitem{Shen}
S.~Shen and R.~Tedrake, ``\BIBforeignlanguage{en}{Sampling {{Quotient}}-{{Ring
  Sum}}-of-{{Squares Programs}} for {{Scalable Verification}} of {{Nonlinear
  Systems}}},'' \emph{\BIBforeignlanguage{en}{MIT CSAIL Public Papers}}, p.~8,
  2020.

\bibitem{Nocedal2006}
J.~Nocedal and S.~J. Wright, \emph{\BIBforeignlanguage{en}{Numerical
  Optimization}}, 2nd~ed., ser. Springer Series in Operations Research.\hskip
  1em plus 0.5em minus 0.4em\relax {New York}: {Springer}, 2006.

\bibitem{Boyd2004}
S.~P. Boyd and L.~Vandenberghe, \emph{\BIBforeignlanguage{en}{Convex
  Optimization}}.\hskip 1em plus 0.5em minus 0.4em\relax {Cambridge, UK ; New
  York}: {Cambridge University Press}, 2004.

\bibitem{Papp2018}
D.~Papp and S.~Y{\i}ld{\i}z, ``\BIBforeignlanguage{en}{Sum-of-squares
  optimization without semidefinite programming},''
  \emph{\BIBforeignlanguage{en}{arXiv:1712.01792 [math]}}, 2018.

\bibitem{Ahmadi2019}
A.~A. Ahmadi and A.~Majumdar, ``\BIBforeignlanguage{en}{{{DSOS}} and {{SDSOS
  Optimization}}: {{More Tractable Alternatives}} to {{Sum}} of {{Squares}} and
  {{Semidefinite Optimization}}},'' \emph{\BIBforeignlanguage{en}{SIAM Journal
  on Applied Algebra and Geometry}}, vol.~3, no.~2, pp. 193--230, 2019.

\bibitem{Singh}
S.~Singh, M.~Chen, S.~L. Herbert, C.~J. Tomlin, and M.~Pavone,
  ``\BIBforeignlanguage{en}{Robust {{Tracking}} with {{Model Mismatch}} for
  {{Fast}} and {{Safe Planning}}: An {{SOS Optimization Approach}}},''
  \emph{\BIBforeignlanguage{en}{International Workshop on the Algorithmic
  Foundations of Robotics}}, p.~20, 2018.

\bibitem{Parrilo2000}
P.~A. Parrilo, ``Structured {{Semidefinite Programs}} and {{Semialgebraic
  Geometry Methods}} in {{Robustness}} and {{Optimization}},'' Ph.D.
  dissertation, California Institute of Technology, 2000.

\bibitem{Meng2020}
F.~Meng, D.~Wang, P.~Yang, G.~Xie, and F.~Guo,
  ``\BIBforeignlanguage{en}{Application of {{Sum}}-of-{{Squares Method}} in
  {{Estimation}} of {{Region}} of {{Attraction}} for {{Nonlinear Polynomial
  Systems}}},'' \emph{\BIBforeignlanguage{en}{IEEE Access}}, vol.~8, pp.
  14\,234--14\,243, 2020.

\bibitem{Majumdar2013}
A.~Majumdar, A.~A. Ahmadi, and R.~Tedrake, ``\BIBforeignlanguage{en}{Control
  design along trajectories with sums of squares programming},'' in
  \emph{\BIBforeignlanguage{en}{2013 {{IEEE International Conference}} on
  {{Robotics}} and {{Automation}}}}.\hskip 1em plus 0.5em minus 0.4em\relax
  {Karlsruhe, Germany}: {IEEE}, 2013, pp. 4054--4061.

\bibitem{Lopez2019}
B.~T. Lopez, J.-J.~E. Slotine, and J.~P. How, ``\BIBforeignlanguage{en}{Dynamic
  {{Tube MPC}} for {{Nonlinear Systems}}},''
  \emph{\BIBforeignlanguage{en}{arXiv:1907.06553 [cs, eess]}}, Jul. 2019.

\bibitem{Li}
H.~X. Li and B.~C. Williams, ``\BIBforeignlanguage{en}{Generative {{Planning}}
  for {{Hybrid Systems}} based on {{Flow Tubes}}},''
  \emph{\BIBforeignlanguage{en}{International Conference on Automated Planning
  and Scheduling}}, p.~8, 2008.

\bibitem{Bansal2017}
S.~Bansal, M.~Chen, S.~Herbert, and C.~J. Tomlin,
  ``\BIBforeignlanguage{en}{Hamilton-{{Jacobi Reachability}}: {{A Brief
  Overview}} and {{Recent Advances}}},''
  \emph{\BIBforeignlanguage{en}{arXiv:1709.07523 [cs, math]}}, Sep. 2017.

\bibitem{Chen2016}
M.~Chen, S.~Herbert, and C.~J. Tomlin, ``\BIBforeignlanguage{en}{Exact and
  {{Efficient Hamilton}}-{{Jacobi}}-based {{Guaranteed Safety Analysis}} via
  {{System Decomposition}}},'' \emph{\BIBforeignlanguage{en}{arXiv:1609.05248
  [math]}}, Sep. 2016.

\bibitem{Cifuentes2017}
D.~Cifuentes and P.~A. Parrilo, ``\BIBforeignlanguage{en}{Sampling {{Algebraic
  Varieties}} for {{Sum}} of {{Squares Programs}}},''
  \emph{\BIBforeignlanguage{en}{SIAM Journal on Optimization}}, vol.~27, no.~4,
  pp. 2381--2404, 2017.

\bibitem{Howell}
T.~A. Howell, B.~E. Jackson, and Z.~Manchester,
  ``\BIBforeignlanguage{en}{{{ALTRO}}: {{A Fast Solver}} for {{Constrained
  Trajectory Optimization}}},'' in \emph{\BIBforeignlanguage{en}{2019
  {{IEEE}}/{{RSJ International Conference}} on {{Intelligent Robots}} and
  {{Systems}} ({{IROS}})}}.\hskip 1em plus 0.5em minus 0.4em\relax {Macau,
  China}: {IEEE}, Nov. 2019, pp. 7674--7679.

\bibitem{Vinh1980}
Vinh, \emph{Hypersonic {{Planetary Entry Flight Mechanics}}}.\hskip 1em plus
  0.5em minus 0.4em\relax University of Michigan Press, 1980.

\bibitem{ManriqueJoelBenito2010}
J.~B. Manrique, ``Advances in {{Spacecraft Atmospheric Reentry Guidance}},''
  Ph.D. dissertation, University of California Irvine, 2010.

\bibitem{Manchester}
Z.~Manchester and S.~Kuindersma, ``\BIBforeignlanguage{en}{{{DIRTREL}}:
  {{Robust Nonlinear Direct Transcription}} with {{Ellipsoidal Disturbances}}
  and {{LQR Feedback}}},'' \emph{\BIBforeignlanguage{en}{Robotics: Science and
  Systems}}, p.~9, 2017.

\bibitem{mosek}
M.~ApS, \emph{{{MOSEK}} Optimization Suite. {{Release}} 9.1.11.}, 2020.

\end{thebibliography}
\thebiography
\vspace{3mm}
\begin{biographywithpic}
{Remy Derollez}{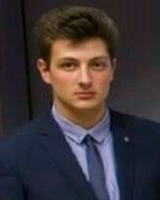}
is a Systems Engineer at Loft Orbital and an alumni of the Robotic Exploration Lab. He receives his BS in Engineering from Ecole Centrale Paris in 2017 and his MS in Aerospace Engineering from Stanford University in 2020. His research interests include dynamical systems, robust motion planning, uncertainty propagation, mathematical methods and optimization for space applications and human space exploration.
\end{biographywithpic} 

\vspace{3mm}
\begin{biographywithpic}
{Simon Le Cleac'h}{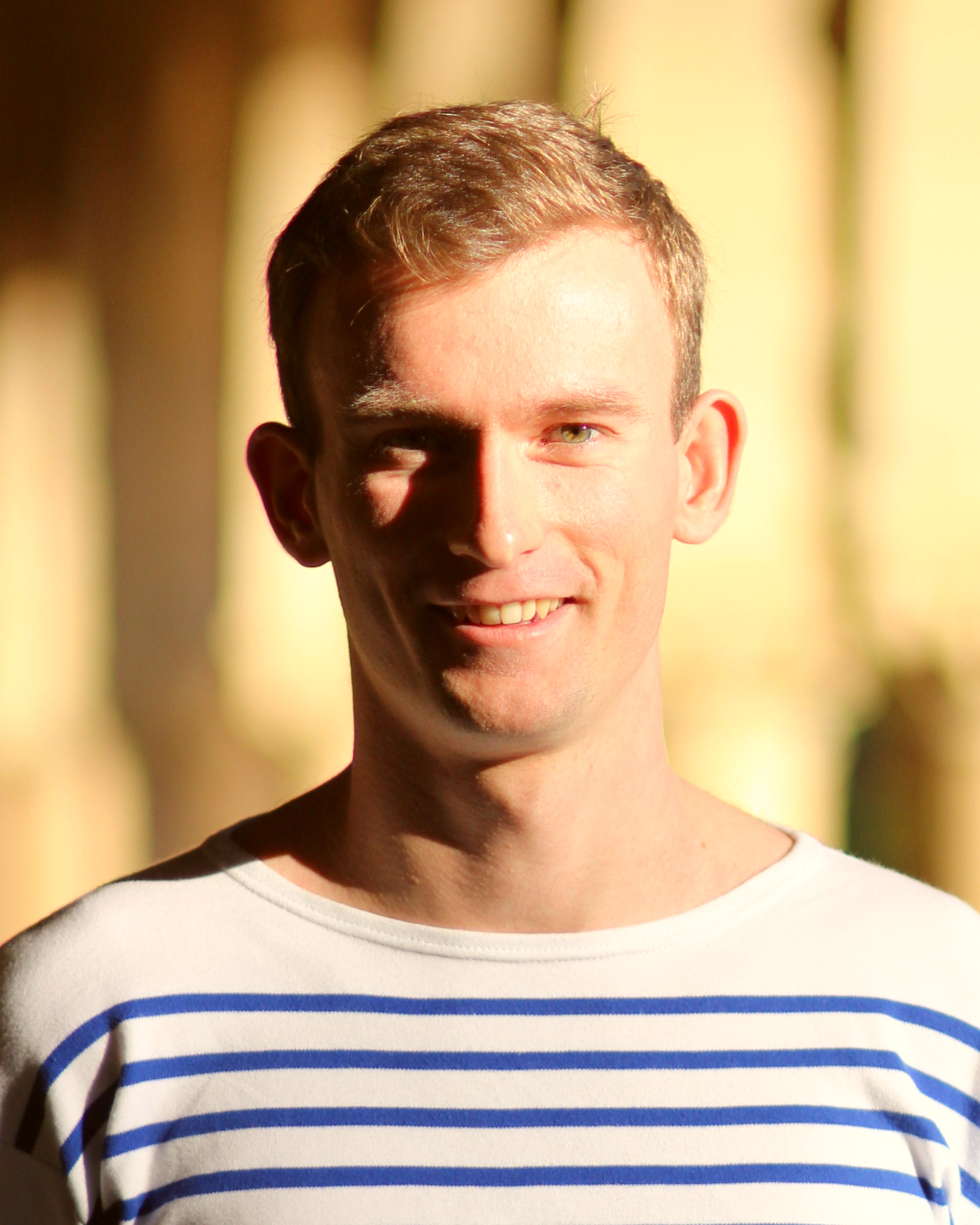}
is a graduate student with the Robotic Exploration Lab at Carnegie Mellon University and with the Multi-Robot Systems Lab at Stanford University. He received his BS in Engineering from Ecole Centrale Paris in 2016 and his MS in Mechanical Engineering from Stanford University in 2019. His research interests include real-time motion-planning, game-theoretic optimization, and inverse optimal control.
\end{biographywithpic}

\vspace{3mm}
\begin{biographywithpic}
{Zac Manchester}{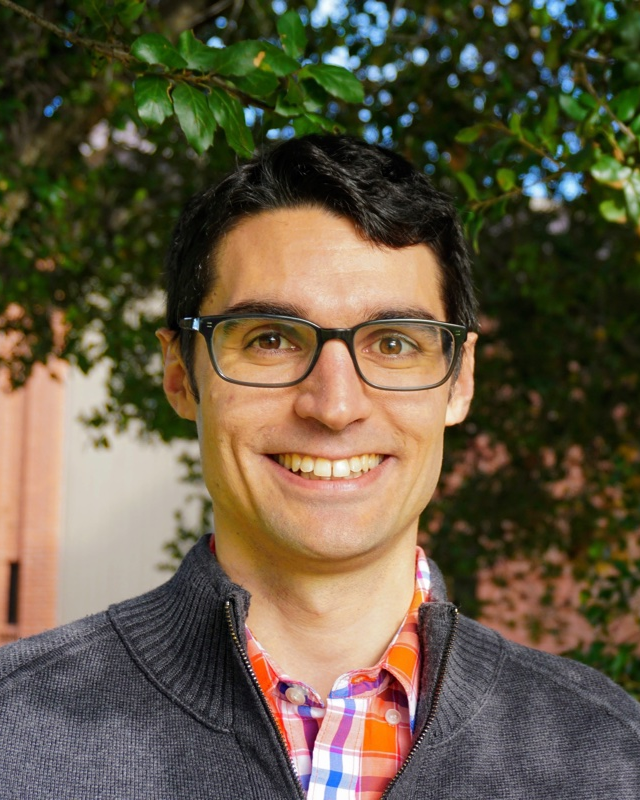}
is an assistant professor in the Robotics Institute at Carnegie Mellon University and founder of the Robotic Exploration Lab. He received a PhD in aerospace engineering in 2015 and a BS in applied physics in 2009, both from Cornell University. His research interest include control and optimization with application to aerospace and robotic systems with challenging nonlinear dynamics.
\end{biographywithpic}

\end{document}